\documentclass[10pt,twocolumn,letterpaper]{article}

\usepackage{cvpr}              
\usepackage[dvipsnames]{xcolor}

\definecolor{cvprblue}{rgb}{0.21,0.49,0.74}
\usepackage[pagebackref,breaklinks,colorlinks,citecolor=cvprblue]{hyperref}
\usepackage{colortbl}
\usepackage{color}
\definecolor{ligntblue}{rgb}{0.8824,1,1}

\usepackage{threeparttable}
\usepackage{algorithm}
\usepackage{algorithmic}

\usepackage{bm}
\usepackage{amsmath}
\usepackage{marvosym}

\usepackage{utfsym}

\newcommand{\format}[2]{\begin{tabular}{@{}c@{}}$#1$\\[-3pt]{\color{gray}$\scriptscriptstyle  \pm  #2$}\end{tabular}}

\newcommand{\blformat}[2]{\color{Blue} \begin{tabular}{@{}c@{}}$\mathbf{#1}$\\[-3pt]$\scriptscriptstyle \mathbf{\pm #2}$\end{tabular}}
\newcommand{\bblformat}[1]{\color{Blue}{\begin{tabular}{@{}c}$\mathbf{#1}$\end{tabular}}}
\newcommand{\fformat}[1]{\color{black}{\begin{tabular}{@{}c}${#1}$\end{tabular}}}
\newcommand{\gR}{\mathcal{R}} 
\newcommand{\sU}{\mathbb{U}} 
\newcommand{\sS}{\mathbb{S}} 
\newcommand{\sB}{\mathbb{B}} 

\makeatletter
\newcommand{\subalign}[1]{%
  \vcenter{%
    \Let@ \restore@math@cr \default@tag
    \baselineskip\fontdimen10 \scriptfont\tw@
    \advance\baselineskip\fontdimen12 \scriptfont\tw@
    \lineskip\thr@@\fontdimen8 \scriptfont\thr@@
    \lineskiplimit\lineskip
    \ialign{\hfil$\m@th\scriptstyle##$&$\m@th\scriptstyle{}##$\hfil\crcr
      #1\crcr
    }%
  }%
}
\makeatother

\DeclareMathOperator*{\argmax}{arg\,max}
\DeclareMathOperator*{\argmin}{arg\,min}
\def\paperID{11387} 
\def\confName{CVPR}
\def\confYear{2024}

\title{Spanning Training Progress: Temporal Dual-Depth Scoring (TDDS) for Enhanced Dataset Pruning}

\author{Xin Zhang\textsuperscript{1,2}\footnotemark[1] \quad
Jiawei Du\textsuperscript{2,3}\footnotemark[2]\quad
Yunsong Li\textsuperscript{1} \quad
Weiying Xie\textsuperscript{1}\footnotemark[3] \quad
Joey Tianyi Zhou\textsuperscript{2,3}\footnotemark[3] \\
\textsuperscript{1}{\small XiDian University, Xi'an, China}\\
\textsuperscript{2}{\small Centre for Frontier AI Research (CFAR), Agency for Science, Technology and Research (A*STAR), Singapore}\\
\textsuperscript{3}{\small Institute of High Performance Computing (IHPC), Agency for Science, Technology and Research (A*STAR), Singapore}\\
\texttt{\small xinzhang01@stu.xidian.edu.cn, \{dujw, Joey\_Zhou\}@cfar.a\-star.edu.sg}\\
\texttt{\small \{ysli@mail,wyxie@\}.xidian.edu.cn}
}

\begin{document}
\maketitle
\renewcommand{\thefootnote}{\fnsymbol{footnote}} 
\footnotetext[2]{Equal contribution.}
\footnotetext[3]{Corresponding authors.}
\footnotetext[1]{Work completed during internship at A*STAR}
\renewcommand{\thefootnote}{\arabic{footnote}} 
\begin{abstract}
Dataset pruning aims to construct a coreset capable of achieving performance comparable to the original, full dataset. Most existing dataset pruning methods rely on snapshot-based criteria to identify representative samples, often resulting in poor generalization across various pruning and cross-architecture scenarios. Recent studies have addressed this issue by expanding the scope of training dynamics considered, including factors such as forgetting event and probability change, typically using an averaging approach. However, these works struggle to integrate a broader range of training dynamics without overlooking well-generalized samples, which may not be sufficiently highlighted in an averaging manner. In this study, we propose a novel dataset pruning method termed as \textbf{Temporal Dual-Depth Scoring (TDDS)}, to tackle this problem. TDDS utilizes a dual-depth strategy to achieve a balance between incorporating extensive training dynamics and identifying representative samples for dataset pruning. In the first depth, we estimate the series of each sample's individual contributions spanning the training progress, ensuring comprehensive integration of training dynamics. In the second depth, we focus on the variability of the sample-wise contributions identified in the first depth to highlight well-generalized samples. Extensive experiments conducted on CIFAR and ImageNet datasets verify the superiority of TDDS over previous SOTA methods. Specifically on CIFAR-100, our method achieves 54.51\% accuracy with only 10\% training data, surpassing baselines methods by more than 12.69\%. 
Our codes are available at \url{https://github.com/zhangxin-xd/Dataset-Pruning-TDDS}.
\end{abstract}    
\section{Introduction}
\label{sec:intro}
Explosively growing datasets \cite{kuznetsova2020open, zhou2017scene, deng2009imagenet,zhou2023dataset} have been crucial for the success of deep neural networks (DNNs) in various applications \cite{liu2021swin, Zhang_2023_ICCV, Cheng_2022_CVPR, Rombach_2022_CVPR}. However, learning from large-scale datasets is both time-consuming and financially prohibitive \cite{strubell2019energy, sorscher2022beyond, dataset2022, nguyen2020dataset,wang2022cafe,songhuaReview23}. Actually, a substantial portion of data samples are redundant \cite{toneva2018an, xia2022moderate, zheng2022coverage}, which means they can be excluded from training without compromising performance.
\begin{figure}
    \centering
    \includegraphics[width=0.96\linewidth]{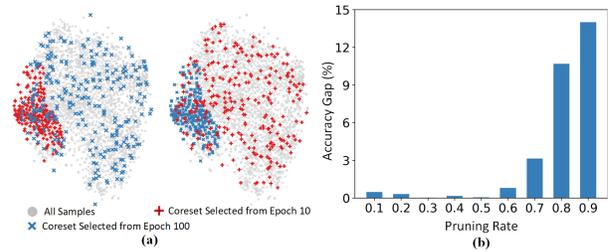}
    \caption
    {(a) Coresets constructed by snapshot-based dataset pruning tend to fluctuate depending on the epochs selected for data selection. The red and blue points represent the data samples of two coresets constructed with 90\% pruning rate at Epoch 10 and 100 from the same training progress, respectively.  (b) Snapshot-based coresets fail to generalize to pruning settings, especially pruning rates. The accuracy gap between coresets constructed at Epoch 10 and 100 is increasing as the pruning rate grows. Here, we measure importance using the error vector score, a classic snapshot-based criterion defined in \cite{paul2021deep}. These scores are obtained by training ResNet-18 on CIFAR-10.}
    \label{fig: start}
\end{figure}

Numerous dataset pruning works \cite{marion2023less, paul2022lottery, xia2022moderate, Sustainable, huang2023efficient, loshchilov2015online, park2023robust,he2024multisize,du2023minimizing} have explored various criteria \cite{guo2022deepcore, tan2023data} for identifying important data samples, including distribution distance \cite{welling2009herding, xia2022moderate}, uncertainty \cite{coleman2019selection}, loss \cite{toneva2018an, paul2021deep}, and gradient \cite{paul2021deep}. 
Most of them evaluate sample importance using a snapshot of training progress. For example, \citet{paul2021deep} uses the error vector score generated by a few epochs\footnote{Usually, the first \(10\) or \(20\) epochs.} in early training; \citet{xia2022moderate} calculates the distribution distances of features at the end of training. 

However, these snapshot-based pruning methods have two potential problems. Firstly, the importance scores of samples fluctuate with epochs during the progress of training. Consequently, the coresets constructed by the importance scores at different snapshots of training progress differ from each other significantly. As verified in \autoref{fig: start}\textcolor{red}{(a)}, the coresets constructed at Epoch 10 and 100 obviously contain different samples. Secondly, relying on importance scores from a snapshot can lead to overfitting in the construction of the coreset, as this approach may not generalize well across different scenarios, such as varying pruning rates. As shown in \autoref{fig: start}\textcolor{red}{(b)}, the accuracy gap of coresets from different snapshots becomes larger as the pruning rate increases. 

To address these two problems, a straightforward approach is to consider a broader range of training dynamics throughout the training process, with a focus on selecting well-generalized data samples. For example, \citet{pleiss2020identifying} involves training dynamics by measuring the probability gap between the target class and the second largest class in each epoch; 
\citet{toneva2018an} monitors the forgetting events throughout the entire training process. These forgetting events occur when samples, classified correctly in a prior epoch, are subsequently predicted incorrectly. Unfortunately, existing studies fall short of balancing the incorporation of training dynamics with the effective identification of well-generalized samples. The use of an averaging down-sampling operation aims to cover a broader range of training dynamics. However, this approach can inadvertently lead to the neglect of well-generalized samples during data selection, ultimately degrading overall performance.

\begin{figure*}[h]
    \centering
    \includegraphics[width=0.85\linewidth]{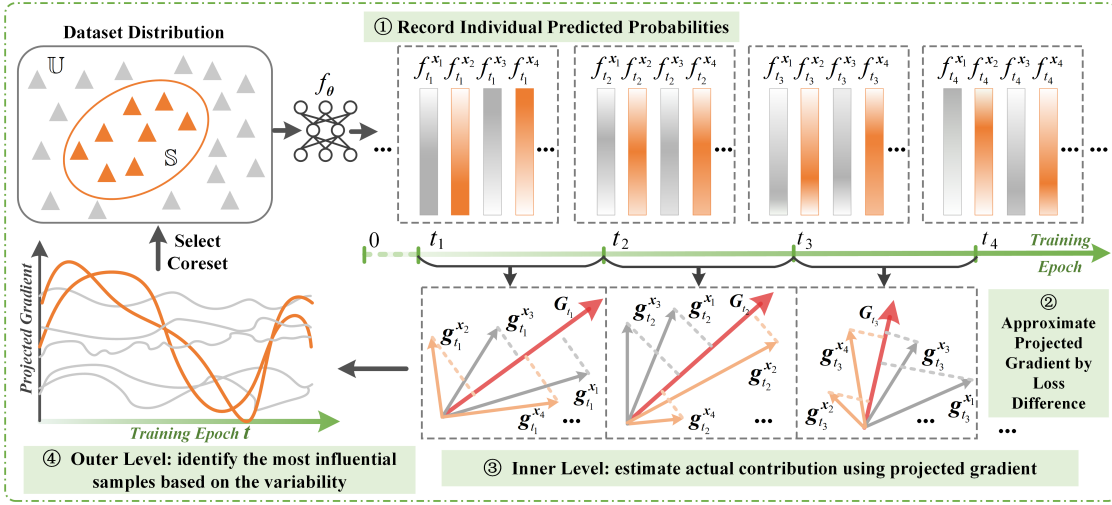}
    \caption{The pipeline of our proposed TDDS. First, we record individual predicted probabilities of samples $f_{t_i}^{x_n} = f_{\bm{\theta}_{t_i}}(\bm{x}_n)$ during training. Then, we approximate the projected gradient, i.e. the component of gradient $\bm{g}_{t_i}^{x_n} = \bm{g}_{t_i}(\bm{x}_n)$ on accumulated gradient $\bm{G}_{t_i}$ direction, with KL-based loss difference between two adjacent epochs. Next, the projected gradients are used to estimate individual contributions of samples. Finally, the important samples are identified with larger projected gradient variances over time.}
    \label{fig:framework}
\end{figure*}

In this paper, we propose \textbf{Temporal Dual-Depth Scoring (TDDS)} pruning method,  designed to effectively balance and address the aforementioned problems. We employ a dual-depth temporal scoring metric that spans the training process for data selection. \textit{In the inner level (depth one)}, we approximate the actual contribution of each sample in every epoch. This approximation is achieved by projecting the sample-wise gradients onto the accumulated gradient in each epoch, following the approach suggested by \citet{du2021efficient}. We then integrate these contributions over a specified window, typically spanning 10 epochs, to incorporate more training dynamics. \textit{In the outer level (depth two)}, we reduce reliance on the averaging down-sampling operation to prevent overlooking vital samples. Well-generalized samples are identified based on the variability of the projected gradients, calculated in each window from the inner level. This design of the outer level draws inspiration from \citet{he2023large}, who tracked the variance of predicted target probabilities throughout training to enhance pruning effectiveness. 
The pipeline of our proposed TDDS is clearly illustrated in \autoref{fig:framework}. Extensive experiments demonstrate the superior generalization capability of the proposed TDDS, enabling it to outperform other comparison methods across different pruning settings on various datasets and networks. 
For example, on CIFAR-100, our method achieves 54.51\% accuracy with only 10\% training data, surpassing random selection by 7.83\% and other SOTA methods by at least 12.69\%.
 
Our main contributions are summarized as: 
\begin{itemize}
\item We examine the poor generalization resulting from existing snapshot-based dataset pruning. Subsequently, we analyze the inefficacy of current solutions to this generalization issue, attributing it to the inability to balance the incorporation of training dynamics with the effective identification of well-generalized samples.
\item We thereby propose a novel dataset pruning method called \textbf{Temporal Dual-Depth Scoring (TDDS)}, resorting to two temporal depths/levels to take respective care of the contribution evaluation and generalization spanning training progress. Experiments on diverse datasets and networks demonstrate the effectiveness of TDDS, achieving SOTA performance.
\end{itemize}

\section{Related Works}
\label{sec:Related Works}
\subsection{Dataset Pruning}
Dataset Pruning, also known as Coreset Selection aims to shrink the dataset scale by selecting important samples according to some predefined criteria. Herding \cite{chen2012super, welling2009herding} and Moderate \cite{xia2022moderate} calculate the distance in feature space. Entropy \cite{coleman2019selection} and Cal \cite{margatina2021active} explore the uncertainty and decision boundary with the predicted outputs. GradND/EL2N \cite{paul2021deep} quantifies the importance of a sample with its gradient magnitude. Most of these methods are snapshot-based, that is they do not care about the training dynamics that actually distinguish the behaviors of samples in the overall optimization.  There are some efficient training methods \cite{tang2023exploring, okanovic2023repeated, xu2023efficient, Adversarial2022}, which focus more on training efficiency and execute online selection spanning training progress. For example, gradient matching-based methods GradMatch \cite{killamsetty2021grad} and Craig \cite{mirzasoleiman2020coresets} minimize the distance between gradients produced by the full dataset and coreset. ACS \cite{huang2023efficient} selects samples with larger gradient magnitude to accelerate quantization-aware training. These methods can capture training dynamics but often necessitate repeated importance re-evaluating and can not be generalized to different networks. Some methods such as Forgetting \cite{toneva2018an} and AUM \cite{pleiss2020identifying} also have proved training dynamics benefits samples importance assessment. However, the downsampling operation, which averages dynamics across training, disregards the variability crucial for evaluating sample generalization, ultimately resulting in suboptimal coresets.
\subsection{Variance in Deep Learning}
Variance, a standard second-order statistic, revealing the variability along a certain dimension, has been explored in numerous works \cite{bardes2021vicreg, pmlr-v119-park20b, rame2022fishr}. For example, MVT \cite{pmlr-v119-park20b} transfers the observed variance of one class to another to improve the robustness of the unseen examples and few-shot learning tasks. VICReg \cite{bardes2021vicreg} maintains variance of embeddings to prevent the collapse in which two branches produce identical outputs regardless of inputs. For the gradient, its variance along the parameter dimension is an approximation of Hessian matrix indicating loss landscape. Based on that, Fishr \cite{rame2022fishr} matches domain-level variances of gradients for out-of-distribution generalization. LCMat \cite{shin2023loss} induces a better adaptation of the pruned dataset on the perturbed parameter region than the exact point matching. Besides, calculating gradient variance along time dimension tracks dynamic optimization change. VoG \cite{agarwal2022estimating} uses this to estimate sample importance and the visualization shows that high-variance samples are more worthwhile to learn. Dyn-Unc \cite{he2023large} calculates the time-dimension variance of predicted target class probability to evaluate sample importance. Differing from VoG and Dyn-Unc, our TDDS does not need pixel-wise gradient calculation while considering non-target probabilities, which enables a more precise evaluation with lower computational overhead.

\section{Preliminary}
Throughout this paper, we denote the full training dataset as $\mathbb{U}=\left\{\left(\bm{x}_n, \bm{y}_n\right)\right\}_{n=1}^N$, where $\bm{x}_n \in \mathbb{R}^D$ and $\bm{y}_n\in \mathbb{R}^{1\times C}$ are drawn i.i.d. from a natural distribution $\mathcal{D}$.
We denote a neural network parameterized with weight matrix $\bm{\theta}$ as $f_{\bm{\theta}}$. Under these definitions, the optimization objective of $f_{\bm{\theta}}$ on $\mathbb{U}$ is to minimize empirical risk $\mathcal{L}(\mathbb{U}; \bm{\theta})=\frac{1}{N} \sum_{n=1}^N \ell\left(f_{\bm{\theta}}(\bm{x}_n),\bm{y}_n\right)$, where $f_{\bm{\theta}}(\bm{x}_n)\in\mathbb{R}^{1\times C}$ outputs the predicted probabilities of each class. Typically, $\bm{\theta}$ is updated as follow\footnote{Here, $t$ indicates epoch and the batch-wise updates \cite{amari1993backpropagation} in one epoch are omitted.},
\begin{equation}\label{SGD}
\bm{\theta}_{t+1}=\bm{\theta}_{t}-\eta \sum_{n=1,\bm{x}_n \in \sU}^{N} \bm{g}_t(\bm{x}_n),
\end{equation}
where $\eta$ is the learning rate and $\bm{g}_t(\bm{x}_n)$ is defined as the gradient calculated over the data pair $(\bm{x}_n, \bm{y}_n)$ in epoch $t$. 

Dataset pruning aims to enhance training efficiency by constructing a coreset $\mathbb{S}=\left\{\left(\bm{x}_n, \bm{y}_m\right)\right\}_{m=1}^M$,  $\mathbb{S}\subset \sU$ that can achieve comparable performance to $\mathbb{U}$. The objective of dataset pruning can be formulated as,
\begin{equation}
\label{Dataset pruning}
\underset{\substack{(\bm{x}, \bm{y}) \sim \mathcal{D} \\ \bm{\theta}_0 \sim \mathcal{P}_{\bm{\theta}_0}}}{\mathbb{E}}\left[\ell\left(f_{\left(\sU, \bm{\theta}_0\right)}(\bm{x}), \bm{y}\right)\right] \simeq \underset{\substack{(\bm{x}, \bm{y}) \sim \mathcal{D} \\ \bm{\theta}_0 \sim \mathcal{P}_{\bm{\theta}_0}}}{\mathbb{E}}\left[\ell\left(f_{\left(\sS, \bm{\theta}_0\right)}(\bm{x}), \bm{y}\right)\right],    
\end{equation}
where \(f_{\left(\sU, \bm{\theta}_0\right)}\) and \(f_{\left(\sS, \bm{\theta}_0\right)}\) represent the networks trained on \(\sU\) and \(\sS\) with weight $\bm{\theta}_0$ initialized from distribution $\mathcal{P}_{\bm{\theta}_0}$.
\section{Methodology}
\label{sec:Methodology}
As discussed earlier, despite that exploring long-range dynamics provides a more comprehensive sample evaluation, the averaging operation hinders the construction of a coreset with strong generalization. In this section, we present the proposed \textbf{Temporal Dual-Depth Scoring (TDDS)} strategy, which balances the incorporation of training dynamics and the identification of well-generalized samples. In the outer level, we identify samples with their variability during training which is encouraged theoretically by the optimization space matching between coreset and original dataset (refer to Section \ref{sec: method_1}). In the inner level, the gradient in the outer level is replaced with the projected component on the accumulated direction to estimate valid contributions in every epoch (refer to Section \ref{sec: method_2}). Finally, to enhance efficiency, we conduct TDDS with a moving window (refer to Section \ref{sec: method_3}).
The proposed TDDS is summarized in \autoref{algo1}. 
\subsection{Outer Level: Enhance Coreset Generalization by Maximizing Variance}\label{sec: method_1}
To construct the coreset $\sS$ satisfying \autoref{Dataset pruning}, gradient-based methods \cite{killamsetty2021grad,mirzasoleiman2020coresets, balles2022gradient} propose to minimize the distance between full dataset and coreset gradient,
\textcolor{black}{
\begin{align}
\label{eq: gradient matching}
\sS^{*} = \argmin_{\sS \subset \sU}& \|\bm{G}_{t,\sU}-\tilde{\bm{G}}_{t,\sS}\|,\quad \mbox{where} \notag \\
\bm{G}_{t,\sU}=\sum_{\substack{n=1, \\ \bm{x}_n \in \sU}}^{|\sU|}  \bm{g}_t(\bm{x}_n), \quad
&\tilde{\bm{G}}_{t,\sS}=\sum_{\substack{m=1, \\\bm{x}_m \in \sS}}^{|\sS|}  \bm{g}_t(\bm{x}_m), 
\end{align}
$\tilde{\bm{G}}_{t,\sS}$ can be regarded as the gradient reconstruction in coreset space.} To solve this minimization problem, \citet{huang2023efficient} prunes samples with smaller gradient magnitudes to minimize their potential impact on optimization.

However, whenever the distance is minimized, the obtained snapshot-based coreset fails to generalize across training progress. Here are the reasons: (1) the distinctive contributions of samples are temporal statistics. Evaluating them at a specific time point causes significant bias when considering dynamic optimization; (2) samples with larger individual gradients dominate the matching process. Without considering the current optimization context, these samples introduce a substantial error in approximating the accumulated gradient.

To address the first issue, we proposed to extend \autoref{eq: gradient matching} to a continuous training duration to capture training dynamics,
\textcolor{black}{
\begin{equation}
\begin{aligned}\label{optimization matching}
\sS^{*} &= \argmin _{\sS \subset \sU} \frac{1}{T}\sum_{t=1}^{T}\|{\bm{\mathcal{G}}}_{t, \sU}-\tilde{\bm{\mathcal{G}}}_{t, \sS}\|^2,
\end{aligned}    
\end{equation}
Here, the gradients generated by $\sU$ at epoch $t$ are defined as $\bm{\mathcal{G}}_{t, \sU}=[|\bm{g}_t(\bm{x}_n)|]_{n=1}^{N}$ and similarly $\tilde{\bm{\mathcal{G}}}_{t, \sS}$ is its reconstructed counterpart using $\sS$. Here we use magnitude instead of gradient itself to balance the training dynamic integration and complexity.
Experiments in Section \ref{sec: Ablation} also shows that magnitude-based minimization performs better than gradient-based when considering continuous training. }

Directly solving \autoref{optimization matching} is NP-hard. Thus, we turn to its equivalent, temporal gradient variance maximization,
\begin{equation}
\begin{aligned}\label{variance}
\mathbb{S}^{*} = \argmax _{\sS \subset \sU} \gR(\sS), 
\quad 
\gR(\sS) = \sum_{t=1}^{T}\|\bm{\mathcal{G}}_{t, \sS}-\overline{\bm{\mathcal{G}}}_{\sS}\|^2,
\end{aligned}    
\end{equation}
\textcolor{black}{where $\overline{\bm{\mathcal{G}}}_{ \sS}=\frac{1}{T}\sum_{t=1}^T\bm{\mathcal{G}}_{t, \sS}$, $\bm{\mathcal{G}}_{t, \sS}=[|\bm{g}_t(\bm{x}_m)|]_{m=1}^{M}$ is the mean gradient over training time.}
Detailed proof can be founded in Supplementary. \autoref{variance} measures the variability of samples and enhances the identification of well-generalized samples compared to previous average-based methods \cite{toneva2018an, pleiss2020identifying}. 
\subsection{Inner Level: Estimate Actual Contribution with Projected Gradient}\label{sec: method_2}
\begin{figure}
    \centering
    \includegraphics[width=1\linewidth]{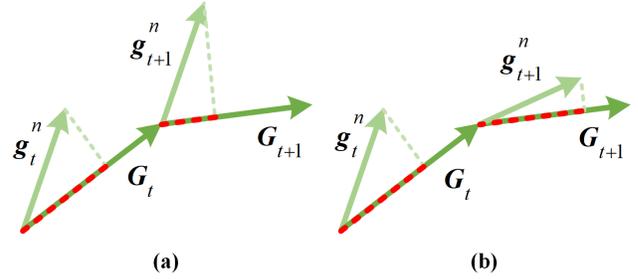}
    \caption{Incorporating sample-wise gradient in the current optimization context. After projecting onto the current accumulated gradient, samples with similar gradients exhibit different behaviors.}
    \label{fig: projection}
\end{figure}
According to \autoref{variance}, sample importance can be evaluated with its gradient variance over time. For any sample $\bm{x}_n$, we have,
\begin{equation}
\begin{aligned}\label{variance_sample}
\gR(\bm{x}_n) = \sum_{t=1}^{T}\||\bm{g}_t(\bm{x}_n)|-\overline{|\bm{g}(\bm{x}_n)|}\|^2,
\end{aligned}    
\end{equation}
where $\overline{|\bm{g}(\bm{x}_n)|}=\frac{1}{T}\sum_{t=1}^T|\bm{g}_t(\bm{x}_n)|$. \autoref{variance_sample} 
treats sample-wise gradient independently, which may not precisely describe the valid contribution in the current optimization context. That is the second problem mentioned in Section \ref{sec: method_1}. As illustrated in \autoref{fig: projection}\textcolor{red}{(a)}, the gradient of sample $\bm{x}_n$ remains constant between $t$ and $t+1$. However, when incorporating it into the current optimization context, its valid contribution undergoes a significant shift. Thus, we propose to calculate the gradient variance after projecting it on the accumulated direction, that is, $|\bm{g}_t(\bm{x}_n)|$ is replaced by,
\begin{equation}\label{projection}
\begin{aligned}
|\bm{g}_t(\bm{x}_n)| &\Leftarrow
|\bm{g}_t(\bm{x}_n)\cdot\sum_{\substack{j=1,\bm{x}_j \in \sU}}^{|\sU|}  \bm{g}_t(\bm{x}_j)|.
\end{aligned}    
\end{equation}

Calculating \autoref{variance_sample} requires sample-wise gradients over a period of $T$, which is not feasible due to time and memory constraints. According to \autoref{SGD}, the accumulated gradient is equal to $\frac{1}{\eta}(\bm{\theta}_t-\bm{\theta}_{t+1})$, thus the projection in \autoref{projection} is reformed as loss difference,
\begin{equation}
\begin{aligned}\label{approximat gradient}
|\bm{g}_t(\bm{x}_n)| = &\frac{1}{\eta}|(\bm{\theta}_{t}-\bm{\theta}_{t+1})\nabla_{\bm{\theta}_{t}}\ell\left(f_{\bm{\theta}_t}(\bm{x}_n), \bm{y}_n \right)|
\\\approx&\frac{1}{\eta}|\ell\left(f_{\bm{\theta}_{t+1}}(\bm{x}_n), \bm{y}_n \right)-\ell\left(f_{\bm{\theta}_{t}}(\bm{x}_n), \bm{y}_n \right)|,
\end{aligned}    
\end{equation}
which can be deduced from the first-order Taylor expansion. 

Thus, we can evaluate the per-sample contribution to the accumulated gradient direction with its loss change. In the context of classification, $\ell$ is typically Cross Entropy (CE) loss,
\begin{equation}\label{R ce}
\begin{aligned}
\Delta\ell_t^n= \bm{y}_n^\top\cdot log\frac{f_{\bm{\theta}_{t+1}}(\bm{x}_n)}{f_{\bm{\theta}_{t}}(\bm{x}_n)},
\end{aligned}
\end{equation}
where $\Delta\ell_t^n = \ell\left(f_{\bm{\theta}_{t+1}}(\bm{x}_n), \bm{y}_n \right)-\ell\left(f_{\bm{\theta}_{t}}(\bm{x}_n), \bm{y}_n \right)$. Due to the one-hot $\bm{y}_n$, non-target probabilities are unfortunately overlooked, obliterating a significant amount of distinctive information. This reminds us of the Kullback-Leibler (KL) divergence loss, which can be achieved by simply replacing $\bm{y}_n$ with $f_{\bm{\theta}_{t+1}}(\bm{x}_n)$,
\begin{equation}\label{loss kl}
\begin{aligned}
\Delta\ell_t^n = f_{\bm{\theta}_{t+1}}(\bm{x}_n)^\top\cdot log\frac{f_{\bm{\theta}_{t+1}}(\bm{x}_n)}{f_{\bm{\theta}_{t}}(\bm{x}_n)}.
\end{aligned}
\end{equation}
Considering that $f_{\bm{\theta}_t(\bm{x}_n)}$ follows a Bernoulli distribution $\mathcal{B}(p_t)$, where $p_t$ represents the probability of the target class at epoch $t$. If $p_t$ equals $p_{t+1}$, \autoref{R ce} fails to distinguish, but \autoref{loss kl} can still differentiate the contributions of samples with $q_t$ and $q_{t+1}$. Here, $q_t$ and $q_{t+1}$ are the non-target probability distributions. This effectiveness has been proved in Section \ref{sec: Ablation}.
\subsection{Efficient Implementation of TDDS}\label{sec: method_3}
When computing $\gR(\bm{x}_n)$ with \autoref{variance_sample}, \autoref{approximat gradient}, and \autoref{loss kl}, storing $T$-epoch predicted outputs may lead to out-of-memory problems, especially with large datasets like ImageNet (containing \(1,281,167\) samples across \(1,000\) classes). As a solution, we suggest an efficient calculation using an exponential moving average (EMA) weighting strategy. For any training point $t\geq K$, the variance ${\gR}_t(\bm{x}_n)$\footnote{The constant coefficient $\eta$ has been omitted.} is computed over a $K$-epoch window ranging from $t-K+1$ to $t$,
\begin{equation}\label{window_variance}
\begin{aligned}
{\gR}_t(\bm{x}_n)&= \sum_{t-K+1}^{t}\||\Delta\ell_t^n|-\overline{|\Delta\ell_t^n|}\|^2,
\end{aligned}
\end{equation}
where $\overline{|\Delta\ell_t^n|}$ is the average of loss difference in a window. For $T$-epoch trajectory, $T-K+1$ windows are weighted as,
\begin{equation}\label{average R ema}
\begin{aligned}
\gR(\bm{x}_n) &=\beta {\gR}_t(\bm{x}_n)+(1-\beta)\gR(\bm{x}_n),
\end{aligned}
\end{equation}
where $\beta$ is the decay coefficient of EMA whose setting is examined in Section \ref{sec: para}. With this EMA, we only need to store predicted outputs for one window, significantly alleviating storage burden.

After sorting $\{\gR(\bm{x}_n)\}_{n=1}^N$, the top-M samples are selected to construct coreset denoted as $\{\mathbb{S}, r \} = \{\left(\bm{x}_m, \bm{y}_m\right), \gR(\bm{x}_m)\}_{m=1}^{M}$. Since, the samples with larger $\gR(\bm{x}_m)$ contribute more in the original data training. We remark that this dominance should be maintained in coreset training, i.e.,
\begin{equation}\label{subset_training}
\begin{aligned}
\bm{\theta}_{t+1}=\bm{\theta}_{t}-\eta \sum_{m=1}^M \bm{g}_t(\bm{x}_m)\gR(\bm{x}_m).
\end{aligned}
\end{equation}
This helps to transfer relative sample importance to coreset training. This effectiveness is proved in Section \ref{sec: Ablation}.
\begin{algorithm}
	\caption{Temporal Dual-Depth Scoring (TDDS)}
	\label{algo1}
 \begin{minipage}{0.99\linewidth}
	  {\bf Input:} Training dataset $\sU$; A network $f_{\bm{\theta}}$ with weight $\bm{\theta}$; Learning rate $\eta$; Epochs $T$; Iterations $I$ per epoch; Window size $K$; Decay factor $\beta$.
	\begin{algorithmic}[1]
            \FOR{$t=1$ to $T$}
            \FOR{$i=1$ to $I$, Sample a mini-batch $\sB_i \subset \sU$}
		\STATE Record predicted probabilities $f_{\bm{\theta}_t}(\bm{x}_n)$, $\bm{x}_n \in \sB_i$
           \STATE Calculate $\Delta\ell_t^n$ for each $\bm{x}_n \in \sB_i$
            \\ \COMMENT{Defined in \autoref{loss kl}} 
            \STATE Update weight $\bm{\theta}_{i+1}=\bm{\theta}_{i}-\eta \bm{G}_{t,\sB_i}$
            \ENDFOR
            \IF{$K\leq t\leq T$}
            \STATE Calculate ${\gR}_{t}(\bm{x}_n)$ in one window, $\bm{x}_n \in \sU$\\
            \COMMENT{Defined in \autoref{window_variance}}
            \STATE Update ${\gR}(\bm{x}_n)$ with EMA, $\bm{x}_n \in \sU$\\
            \COMMENT{Defined in \autoref{average R ema}}
            \ENDIF
            \ENDFOR
		\STATE Select Top-M $\{\mathbb{S}, r \}\leftarrow\{\left(\bm{x}_m, \bm{y}_m\right), {\gR}(\bm{x}_m)\}_{m=1}^{M}$
	\end{algorithmic}
 \end{minipage}
	\hspace*{0.02in} {\bf Output:} Selected subset $\mathbb{S}=\left\{\left(\bm{x}_m, \bm{y}_m\right)\right\}_{m=1}^M$; associated sample-wise importance $r=\{{\gR}(\bm{x}_m)\}_{m=1}^{M}$
\end{algorithm}
\section{Experiments}
\begin{table*}[h]
\renewcommand{\arraystretch}{0.9}
	\caption{Comparison of different dataset pruning methods on CIFAR-10 and 100 with ResNet-18 under Strategy-E. The model trained with the full dataset achieves \(95.23\%\) and \(78.21\%\) accuracy.}
	\label{Strategy-E cifar}
	\begin{center}
		\begin{small}
			\setlength{\tabcolsep}{1.8mm}
			\begin{tabular}{cccccccccccc}
				\toprule
                    & \multicolumn{5}{c}{\textbf{CIFAR-10}}&\multicolumn{5}{c}{\textbf{CIFAR-100}}\\ \cmidrule(r){2-6} \cmidrule(r){7-11}
$p$&\fformat{30\%}&\fformat{50\%}&\fformat{70\%}&80\%&\fformat{90\%}&30\%&\fformat{50\%}&\fformat{70\%}&\fformat{80\%}&\fformat{90\%}
\\ \cmidrule(r){1-1}\cmidrule(r){2-6} \cmidrule(r){7-11}
Random  &\format{94.58}{0.04}&\format{93.38}{0.17}&\format{90.61}{0.44}&\format{88.87}{0.47}&\format{83.77}{0.26}&\format{75.53}{0.04}&\format{71.95}{0.16}&\format{64.59}{0.32}&\format{57.79}{0.24}&\format{46.68}{1.07}
\\Entropy \cite{coleman2019selection}    &\format{94.45}{0.07}&\format{91.90}{0.16}&\format{86.24}{0.26}&\format{83.49}{0.21}&\format{72.06}{0.81}&\format{72.39}{0.20}&\format{64.44}{0.36}&\format{50.73}{0.86}&\format{42.86}{0.25}&\format{29.56}{0.54} \\	
Forgetting \cite{toneva2018an}    &\format{95.45}{0.24}&\format{95.05}{0.05}&\format{89.14}{2.04}&\format{76.18}{3.18}&\format{45.87}{1.87}&\format{77.38}{0.09}&\format{70.76}{0.40}&\format{49.92}{0.28}&\format{38.42}{1.13}&\format{25.82}{0.52}\\		
EL2N \cite{paul2021deep}     &\format{ 95.43}{0.10}&\format{95.06}{0.04}&\format{86.69}{1.71}&\format{68.64}{3.70}&\format{31.89}{1.51}&\format{76.89}{0.31}&\format{67.57}{0.15}&\format{36.45}{1.36}&\format{17.31}{0.33}&\format{9.10}{0.69}\\
AUM \cite{pleiss2020identifying}    &\format{95.44}{0.09} &\format{95.19}{0.09}&\format{91.19}{0.63} &\format{69.60}{3.11}&\format{
34.74}{0.11}&\format{77.35}{0.18}&\format{68.17}{0.52}&\format{31.69}{0.34}&\format{18.43}{0.47}
&\format{9.29}{0.27}\\
				Moderate \cite{xia2022moderate}    &\format{93.96}{0.06} &\format{92.34}{0.09} &\format{89.71}{0.14}&\format{
87.75}{0.27}&\format{
83.61}{0.24}&\format{74.60}{0.10}&\format{70.29}{0.31}&\format{62.81}{0.08}&\format{56.52}{0.37}&\format{41.82}{1.12} \\
Dyn-Unc\cite{he2023large}&\format{95.08}{0.02}
&\format{94.03}{0.14}&\format{89.40}{0.13}&\format{79.76}{1.09}&\format{37.12}{1.12}&\format{73.36}{0.10}
&\format{65.90}{0.25}&\format{50.16}{0.47}&\format{39.19}{0.27}&\format{15.20}{0.41}
\\
TDDS&\blformat{95.47}{0.06}&\blformat{95.21}{0.04}&\blformat{93.03}{0.25}&\blformat{91.30}{0.25}&\blformat{85.46}{0.21}&\blformat{77.56}{0.06}&\blformat{74.04}{0.34}&\blformat{67.78}{0.44}&\blformat{63.01}{0.12}&\blformat{54.51}{0.22}\\
				\bottomrule
			\end{tabular}
		\end{small}
  \end{center}
\end{table*}
\begin{table}[h]
	\caption{Comparison of different dataset pruning methods on CIFAR-10 and 100 with ResNet-18 under Strategy-P. The model trained with the full dataset achieves \(95.23\%\) and \(78.21\%\) accuracy.}
	\label{Strategy-P cifar}
	\begin{center}
			\setlength{\tabcolsep}{6pt} 
\renewcommand{\arraystretch}{0.9} 
\resizebox{0.48\textwidth}{!}{
			\begin{tabular}{ccccccc}
				\toprule
                    & \multicolumn{3}{c}{\textbf{CIFAR-10}}& \multicolumn{3}{c}{\textbf{CIFAR-100}}\\ \cmidrule(r){2-4} \cmidrule(r){5-7}
	$p$&30\%&\fformat{50\%}&\fformat{70\%}&30\%&\fformat{50\%}&\fformat{70\%}
\\ \cmidrule(r){1-1}\cmidrule(r){2-4} \cmidrule(r){5-7}
Random  & \format{94.33}{0.17}&\format{93.4}{0.17}&\format{90.94}{0.38}& \format{74.59}{0.27}&\format{71.07}{0.4}&\format{65.3}{0.21}\\
				Entropy \cite{coleman2019selection}&\format{94.44}{0.2}&\format{92.11}{0.47}&\format{85.67}{0.71}&\format{72.26}{0.08}&\format{63.26}{0.29}&\format{50.49}{0.88}
\\	
    Forgetting \cite{toneva2018an}&\format{95.36}{0.13}&\format{95.29}{0.18}&\format{90.56}{1.8}&\format{76.91}{0.32}&\format{68.6}{1.02}&\format{38.06}{1.14}

 \\	
				 EL2N \cite{paul2021deep}&   \format{95.44}{0.15}&\format{94.61}{0.20}&\format{87.48}{1.33}&\format{76.25}{0.24}&\format{65.90}{1.06}&\format{34.42}{1.50}

\\
				AUM \cite{pleiss2020identifying}& \format{95.07}{0.24}&\format{95.26}{0.15}&\format{91.36}{1.4}&\format{76.93}{0.32}&\format{67.42}{0.49}&\format{30.64}{0.58}
 \\
				Moderate \cite{xia2022moderate}&\format{93.86}{0.11}&\format{92.58}{0.30}&\format{90.56}{0.27}&\format{74.60}{0.41}&\format{71.10}{0.24}&\format{65.34}{0.41}
\\
				CCS \cite{zheng2022coverage} &\format{95.40}{0.12}&\format{95.04}{0.37}&\format{93.00}{0.16}&\format{77.14}{0.31}&\format{74.45}{0.16}&\format{68.92}{0.12}\\	 
   \textbf{TDDS}&\blformat{95.50}{0.07}&\blformat{95.66}{0.08}&\blformat{93.92}{0.04}&\blformat{77.95}{0.30}&\blformat{74.98}{0.16}&\blformat{69.46}{0.10}\\ \bottomrule
\end{tabular}
}
\end{center}
\end{table}
\begin{table}[h]
	\caption{Comparison of different dataset pruning methods on ImageNet-1K with ResNet-34 under Strategy-E. The model trained with the full dataset achieves \(73.54\%\) Top-1 accuracy.}
	\label{ImageNet}
	\begin{center}
			\setlength{\tabcolsep}{10pt} 
\renewcommand{\arraystretch}{1.1} 
\resizebox{0.5\textwidth}{!}{
			\begin{tabular}{cccc}
				\toprule
                    & \multicolumn{3}{c}{\textbf{ImageNet-1K}}\\ \cmidrule(r){2-4} 
				 $p$&\fformat{70\%}&\fformat{80\%}&\fformat{90\%}
\\ \cmidrule(r){1-1}\cmidrule(r){2-4} 
Random  &\fformat{64.19}&\fformat{60.76}&\fformat{52.63}\\
Entropy\cite{coleman2019selection}&\fformat{62.34}&\fformat{56.80}&\fformat{43.39}\\	
Forgetting \cite{toneva2018an}&\fformat{64.29}&\fformat{62.01}&\fformat{52.14}\\	
EL2N \cite{paul2021deep}&\fformat{46.92}&\fformat{32.68}&\fformat{15.90}\\
AUM \cite{pleiss2020identifying}
&\fformat{39.34}&\fformat{23.64}&\fformat{11.70}\\
Moderate \cite{xia2022moderate}&\fformat{64.04} &\fformat{61.35}&\fformat{52.45}\\	
\textbf{TDDS}&\bblformat{64.69}&\bblformat{62.56}&\bblformat{53.91}\\ \bottomrule
			\end{tabular}
   }
	\end{center}
\end{table}
\begin{figure*}[t]
    \centering
    \includegraphics[width=1\linewidth]{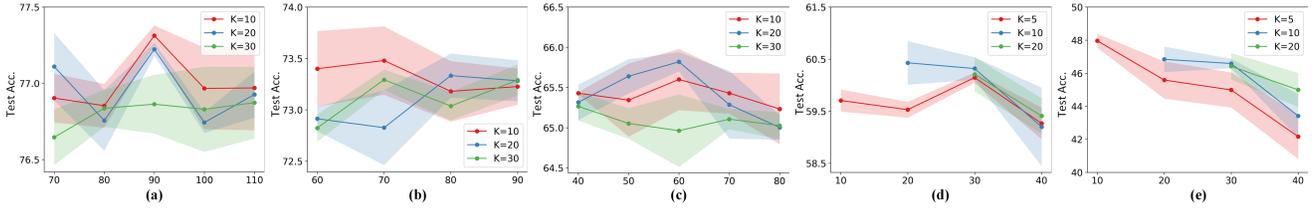}
    \caption{Parameter analysis of range $T$ and window size $K$ on CIFAR-100 with ResNet-18. From left to right, the corresponding pruning rates are $0.3$, $0.5$, $0.7$, $0.8$, and $0.9$.}
    \label{fig:para-100}
\end{figure*}
\begin{figure}[t]
    \centering
    \includegraphics[width=1\linewidth]{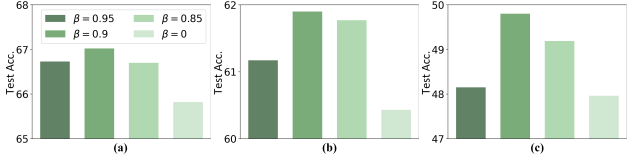}
    \caption{Parameter analysis of decay coefficient $\beta$ on CIFAR-100 with ResNet-18. From left to right, the corresponding pruning rates are \(0.7\), \(0.8\), and \(0.9\).}
    \label{fig:EMA_parameter}
\end{figure}
\subsection{Experiment Settings} 
\textbf{Datasets and Networks.} The effectiveness of the proposed TDDS is evaluated on three popular benchmarks, i.e., CIFAR-10, CIFAR-100 \citep{krizhevsky2009learning}, and ImageNet-1K \cite{deng2009imagenet} with three networks, i.e., ResNet-18, ResNet-34 \citep{he2016deep}, and Swin-T \cite{liu2021swin}. We also verify the cross-architecture performance on four networks including ResNet-50 \citep{he2016deep}, VGG-16 \cite{vgg}, MobileNet-v2 \cite{sandler2018mobilenetv2}, and ShuffleNet \cite{ma2018shufflenet}.\\
\textbf{Training Hyperparameters.} Theoretically, if pruning rate \(p={1-\frac{|\sS|}{|\sU|}}\), where \(0<p \leq1\), the training process speeds up by \(\frac{1}{1-p}\) times. We consider two coreset training strategies: \textbf{Strategy-E} prioritizes efficiency with same-epoch training on both full dataset and coreset. \textbf{Strategy-P} prioritizes performance by training coreset for more epochs. If \(\frac{|\sS|}{|\sU|}=1-p\), the training process is extended to \(\frac{1}{1-p}\) times of the original one. For CIFAR-10 and -100, we train ResNet-18 for \(200\) epochs with a batch size of \(128\). For ImageNet, ResNet-34 and Swin-T are trained for \(60\) \cite{zheng2022coverage} and \(300\) epochs \cite{he2023large} with batch sizes of \(256\) and \(1024\), respectively. SGD optimizer with momentum of \(0.9\) and weight decay of 0.0005 is used for update networks. The learning rate is initialized as \(0.1\) and decays with the cosine annealing scheduler. During experiments, we found that smaller batch size boosts performance at high pruning rates (shown in Supplementary). Thus, for \(p=0.8\), we halved the batch size, and for \(p=0.9\), we reduced it to one-fourth. All compared methods use the same settings. With Strategy-P, for example, training a CIFAR coreset with only \(50\%\)  training samples takes \(400\) epochs, while keeping the other hyperparameters unchanged. Considering the time-consuming training, we only conduct Strategy-P experiments at pruning rates of \(0.3\), \(0.5\), \(0.7\). \\ 
\noindent\textbf{Baselines.}
Eight baselines are used for comparison, with the latter seven being SOTA methods: 1) \textbf{Random}; 2) \textbf{Entropy} \cite{coleman2019selection}; 3) \textbf{Forgetting} \cite{toneva2018an}; 4) \textbf{EL2N} \cite{paul2021deep}; 5) \textbf{AUM} \cite{pleiss2020identifying}; 6) \textbf{Moderate} \cite{xia2022moderate}; 7) \textbf{CCS} \cite{zheng2022coverage}; 8) \textbf{Dyn-Unc} \cite{he2023large}. Due to the limited page, we provide the technical details of these baselines in Supplementary.
\subsection{Benchmark Evaluation Results}
\textbf{ResNet-18 on CIFAR-10 and 100.} \autoref{Strategy-E cifar} and \autoref{Strategy-P cifar} report the performance of different dataset pruning methods under Strategy-E and P, respectively. Under both strategies, TDDS consistently outperforms other methods. For instance, under Strategy-E, none of the compared methods surpasses Random at aggressive pruning rates, while TDDS achieves \(85.46\%\) and \(54.51\%\) accuracy after pruning \(90\%\) samples of CIFAR-10 and 100 surpassing Random by \(1.69\%\) and \(7.83\%\), respectively. By extending the training duration, Strategy-P enhances the performance of all methods to varying degrees. TDDS still maintains its superiority, for example, with only \(30\%\) training data, the accuracy on CIFAR-10 and -100 are \(93.92\%\) and \(69.46\%\) outperforming CCS by \(0.92\%\) and \(0.54\%\), respectively. These results highlight TDDS as a promising choice for both efficiency and performance-oriented dataset pruning. 
\\\textbf{ResNet on ImageNet-1K.} For ImageNet-1K, we train ResNet-34 models on original dataset for \(90\) epochs to collect training dynamics and then train the model on coreset for \(60\) epochs as \cite{zheng2022coverage}. Experiment results are given in \autoref{ImageNet}. The results show that the proposed TDDS has better performance on ImageNet-1K, especially under aggressive pruning rates. For example, when pruning \(90\%\) samples, TDDS still achieves \(53.91\%\) which is 1.77\% and 1.46\% higher than Forgetting and Moderate, respectively.
\\\textbf{Swin-T on ImageNet-1k.} We also evaluate Swin-T on ImageNet-1K. Like Dyn-Unc \cite{he2023large}, \autoref{ImageNet Swin-T} reports the results under four pruning rates. Notably, TDDS consistently outperforms other competitors, for example, after pruning 50\% training data, TDDS achieves 78.33\% accuracy, surpassing Forgetting, Moderate, and Dyn-Unc by 4.01\%, 3.35\% and 0.69\%, respectively. 
\begin{table}[ht]
\renewcommand{\arraystretch}{1}
	\caption{Comparison of different dataset pruning methods on ImageNet-1K with Swin-T under Strategy-E. The model trained with the full dataset achieves \(79.58\%\) Top-1 accuracy.}
 \vspace{-1em}
	\label{ImageNet Swin-T}
	\begin{center}
			\setlength{\tabcolsep}{6pt} 
\renewcommand{\arraystretch}{1.2} 
\resizebox{0.5\textwidth}{!}{
			\begin{tabular}{ccccc}
				\toprule
                    & \multicolumn{4}{c}{\textbf{ImageNet-1K}}\\ \cmidrule(r){2-5} 
$p$&\fformat{25\%}&\fformat{30\%}&\fformat{40\%}&\fformat{50\%}
\\ \cmidrule(r){1-1}\cmidrule(r){2-5} 
Random  &\fformat{77.82}&\fformat{77.18}&\fformat{75.93}&\fformat{74.54}\\
Forgetting \cite{toneva2018an}&\fformat{78.70}&\fformat{78.27}&\fformat{77.55}&\fformat{74.32}\\	
Moderate \cite{xia2022moderate}&\fformat{77.74}&\fformat{77.06}&\fformat{75.94}&\fformat{74.98}\\
Dyn-Unc \cite{he2023large}&\fformat{79.54}&\fformat{79.14}&\fformat{78.49}&\fformat{77.64}\\
\textbf{TDDS}& \bblformat{79.76}&\bblformat{79.57}&\bblformat{79.09}&\bblformat{78.33}\\
						\bottomrule
			\end{tabular}
   }
	\end{center}
\end{table}
\subsection{Parameter Analysis}\label{sec: para}
\begin{figure*}
    \centering
    \includegraphics[width=1\linewidth]{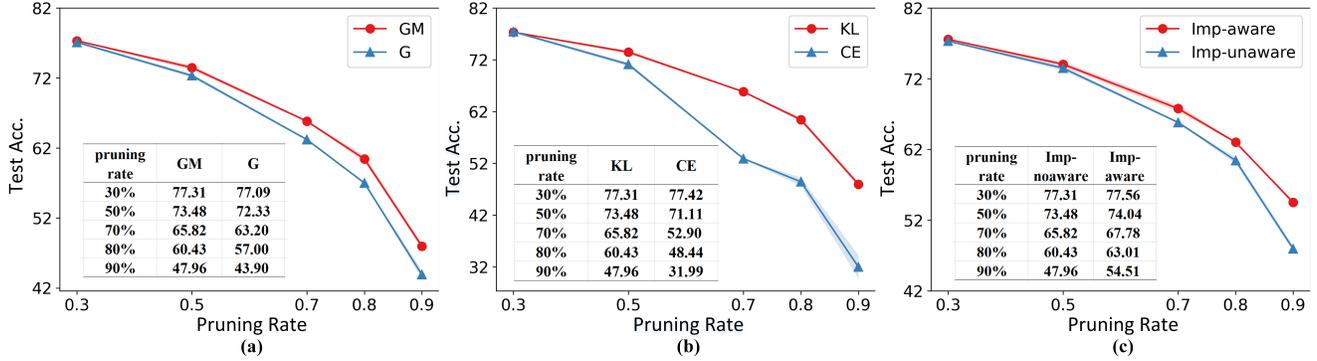}
    \caption{Ablation studies for (a) Gradient Magnitude vs. Gradient, (b) KL-based Loss Difference, and (c) Importance-aware Coreset Training. In (a), GM and G represent estimating sample contribution with gradient magnitude and gradient, respectively. In (b), KL and CE represent calculating loss difference with KL and CE loss, respectively. In (c), Imp-aware and Imp-unaware represent training coreset with and without weighting samples with their importance scores, respectively.}
    \label{fig:ablation_study}
    \vspace{-1em}
\end{figure*}

\begin{table}
    \centering
    \caption{Ablation study for Dual-depth strategy. The numbers in square brackets indicate the improvement in accuracy of the method presented in the current column compared to the corresponding method in the previous column.}
    \begin{tabular}{cccc}
        \toprule
        & \multicolumn{3}{c}{\textbf{CIFAR-100}}\\ \cmidrule(r){2-4} 
        $p$ & EL2N & EL2N+outer &TDDS \\ \hline
        30\% & 76.89& 77.47[\textcolor{teal}{+\phantom{0}0.58}]& \textcolor{Blue}{77.56}[\textcolor{teal}{+\phantom{0}0.09}] \\ 
        50\% & 67.57 & 73.93[\textcolor{teal}{+\phantom{0}6.36}] & \textcolor{Blue}{74.04}[\textcolor{teal}{+\phantom{0}0.11}]\\ 
        70\% & 36.45& 67.04[\textcolor{teal}{+30.59}]& \textcolor{Blue}{67.78}[\textcolor{teal}{+\phantom{0}0.74}]\\ 
        80\% & 17.31 & 58.21[\textcolor{teal}{+40.90}]&\textcolor{Blue}{63.01}[\textcolor{teal}{+\phantom{0}1.80}]\\ 
        90\% & 9.10& 40.36[\textcolor{teal}{+31.26}]& \textcolor{Blue}{54.51}[\textcolor{teal}{+14.15}]\\ \hline
        outer &\usym{2717} & \usym{2713}& \textcolor{Blue}{\usym{2713}}\\ 
        inner &\usym{2717}&\usym{2717}& \textcolor{Blue}{\usym{2713}} \\ \hline
    \end{tabular}
    \label{tab:Dual-Depth}
\end{table}

\textbf{Estimation Range $T$ and Window Size $K$.} Due to the random initialization, models adapt to prominent features (easy samples) during early training. As training progresses, models gradually acquire more abstract and complex features (hard samples) that are challenging to capture in early training. Thus, the importance of samples should be estimated in different training phases with respect to different pruning levels.
As shown in \autoref{fig:para-100}, we analyze the influence of range $T$ and window size $K$ on performance. Apparently, for a larger pruning rate, we should select samples earlier in training. For CIFAR-100, the optimal ($p, T, K$) setting is $(0.3, 90, 10)$, $(0.5, 70, 10)$, $(0.7, 60, 20)$, $(0.8, 20, 10)$, and $(0.9, 10, 5)$. The parameter settings of CIFAR-10 and ImageNet are reported in Supplementary.
\\\textbf{Decay Coefficient $\beta$ of EMA.}  In contrast to the simple average (SA), EMA employs exponential weighting, giving more weight to the most recent window. We examine four values of $\beta$, where $\beta = 0$ represents SA. As illustrated in \autoref{fig:EMA_parameter}, EMA outperforms SA regardless of the $\beta$ value chosen. Among the considered pruning levels, $\beta = 0.9$ yields the best performance, surpassing $\beta = 0.95$ and $\beta = 0.85$. 
\subsection{Ablation Studies}\label{sec: Ablation}
\textbf{Gradient Magnitude vs Gradient.} We use gradient magnitudes in \autoref{optimization matching}. Although gradients themselves appear capable of describing optimization, our experimental findings indicate that the use of gradient magnitudes yields superior performance, as depicted in \autoref{fig:ablation_study}\textcolor{red}{(a)}. We argue that gradients often prioritize samples with frequently reversing gradients, resulting in unstable training and slower convergence.
\\\textbf{KL-based Loss Difference.} As discussed in Section \ref{sec: method_3}, the KL loss provides a more comprehensive assessment of loss difference and is expected to precisely characterize sample contribution. To evaluate its effectiveness, we replace KL loss with CE loss and conduct experiments across all considered pruning rates. As illustrated in \autoref{fig:ablation_study}\textcolor{red}{(b)}, although CE is slightly higher than KL when the pruning rate is 30\%, KL exhibits an absolute advantage at all other pruning levels. For example, when pruning rate is 90\%, KL outperforms CE by a substantial margin of 15.97\%.
\\\textbf{Importance-aware Coreset Training.} We formulate importance-aware subset training as \autoref{subset_training}. The maintained relative importance provides optimization guidance for subset training. As shown in \autoref{fig:ablation_study}\textcolor{red}{(c)}, the advantage becomes increasingly pronounced as the pruning rate rises. When pruning rate is 90\%, the accuracy of importance-aware subset training outperforms its unaware counterpart by 6.55\%.
\\\textbf{Dual-Depth Strategy.}
To investigate the effectiveness of the proposed Dual-Depth strategy, we replace the inner depth with EL2N, and \autoref{tab:Dual-Depth} indicates that incorporating the outer depth boosts EL2N at all considered pruning rates. Nevertheless, this enhancement still falls short of our dual-depth design, particularly evident in aggressive pruning scenarios.

\section{Conclusion}
In this study, we reveal that the snapshot-based data pruning results in poor generalization across various pruning scenarios. However, existing methods fail to identify well-generalized samples even with evaluation criteria considering the whole training progress. We attribute this to the averaging-based dynamics incorporation and imprecise sample contribution characterization. To address these problems, we propose a dataset pruning method called TDDS, which designs a Temporal Dual-Depth Scoring strategy to balance the incorporation of training dynamics and the identification of well-generalized samples. Extensive experiments on various datasets and networks, demonstrate the SOTA performance of our proposed method.

\section*{Acknowledgements}
This research is supported in part by the National Natural
Science Foundation of China under Grant 62121001, Grant 62322117, and Grant 62371365.
This research is also supported by Joey Tianyi Zhou's A*STAR SERC Central Research Fund (Use-inspired Basic Research) and Jiawei Du's A*STAR Career Development Fund (CDF) C233312004.

\clearpage
{
    \small
    \bibliographystyle{ieeenat_fullname}
    \bibliography{main}
}

\clearpage
\setcounter{page}{1}
\maketitlesupplementary
\appendix
\section{Derivation of \autoref{variance}}
\label{MSE to variance}
The objective in \autoref{optimization matching} is to minimize the Mean Squared Error (MSE) between $\bm{\mathcal{G}}_{t, \sU}\in\mathbb{R}^{1\times N}$ and $\tilde{\bm{\mathcal{G}}}_{t, \sS}\in\mathbb{R}^{1\times N}$,
\begin{equation}\label{MSE}
\begin{aligned}
\mathcal{J}= \frac{1}{T}\sum_{t=1}^{T}\|\bm{\mathcal{G}}_{t, \sU}-\tilde{\bm{\mathcal{G}}}_{t, \sS}\|^2.
\end{aligned}    
\end{equation}
Assuming we have a complete $N$-dimension orthonormal basis,
\begin{equation}
\bm{w}_n^T \bm{w}_m=\delta_{n m}= \begin{cases}1, & n=m \\ 0, & \text { else }\end{cases},
\end{equation}
where $\delta_{n m}$ is the kronecker delta and $n,m=1, 2, \cdots N$. Given that any vector can be represented as a linear combination of the basis vectors,
\begin{equation}
\bm{\mathcal{G}}_{t, \sU} = \sum_{n=1}^{N}\alpha_{tn}\bm{w}_n.
\end{equation}
According to the property of orthonormal basis, we have,
\begin{equation}\label{high}
\begin{aligned}
& \alpha_{t n}=\bm{\mathcal{G}}^\top_{t, \sU}\cdot \bm{w}_n \\
&\bm{\mathcal{G}}_{t, \sU} =\sum_{n=1}^N\left(\bm{\mathcal{G}}^T_{t, \sU}\cdot \bm{w}_n\right) \bm{w}_n.
\end{aligned}
\end{equation}
Our goal is to find an $M$-dimension representation $\bm{\mathcal{G}}_{t, \sS}$, 
\begin{equation}\label{low}
\begin{aligned}
\tilde{\bm{\mathcal{G}}}_{t, \sS}
& =\sum_{n=1}^M \alpha_{t n} \bm{w}_n+\sum_{n=M+1}^N {b}_n \bm{w}_n.
\end{aligned}
\end{equation}
The second term indicates bias. With \autoref{high} and \autoref{low}, we can calculate the difference between $\bm{\mathcal{G}}_{t, \sU}$ and $\tilde{\bm{\mathcal{G}}}_{t, \sS}$,
\begin{equation}\label{Difference}
\begin{aligned}
\bm{\mathcal{G}}_{t, \sU}-\tilde{\bm{\mathcal{G}}}_{t, \sS}=&\sum_{n=1}^N\alpha_{t n}\bm{w}_n-\sum_{n=1}^M\alpha_{t n}\bm{w}_n -\sum_{n=M+1}^N {b}_n \bm{w}_n\\
=&\sum_{n=M+1}^N\alpha_{t n}\bm{w}_n- {b}_n \bm{w}_n.
\end{aligned}
\end{equation}
After substituting \autoref{Difference} in \autoref{MSE}, we have 
\begin{equation}\label{MSE_1}
\begin{aligned}
\mathcal{J}= \frac{1}{T}\sum_{t=1}^{T}\|\sum_{n=M+1}^N\alpha_{t n}\bm{w}_n- {b}_n \bm{w}_n\|^2.
\end{aligned}    
\end{equation}
Taking derivative \textit{w.r.t} $\alpha_{t n}$ and ${b}_n$ and setting to zero, we have,
\begin{equation}
\begin{aligned}
{b}_n = \overline{\bm{\mathcal{G}}}^\top_{ \sU}\cdot\bm{w}_n,
\end{aligned}    
\end{equation}
where $\overline{\bm{\mathcal{G}}}_{ \sU}=\frac{1}{T}\sum_{t=1}^T\bm{\mathcal{G}}_{t, \sU}$.
Thus, \autoref{MSE_1} can be reformed as 
\begin{equation}
\begin{aligned}
\mathcal{J}  =&\frac{1}{T} \sum_{t=1}^T\left\|\sum_{n=M+1}^N\left(\bm{\mathcal{G}}^\top_{t, \sU}\cdot\bm{w}_n-\overline{\bm{\mathcal{G}}}^\top_{ \sU}\cdot\bm{w}_n\right) \bm{w}_n\right\|^2 \\
 =&\frac{1}{T} \sum_{t=1}^T\left\|\sum_{n=M+1}^N\left((\bm{\mathcal{G}}_{t, \sU}-\overline{\bm{\mathcal{G}}}_{ \sU})^\top\cdot\bm{w}_n\right)\bm{w}_n\right\|^2  \\
=& \frac{1}{T} \sum_{t=1}^T(\sum_{n=M+1}^N\left((\bm{\mathcal{G}}_{t, \sU}-\overline{\bm{\mathcal{G}}}_{ \sU})^\top\cdot\bm{w}_n\right) \bm{w}_n)^\top\cdot\\&(\sum_{n=M+1}^N\left((\bm{\mathcal{G}}_{t, \sU}-\overline{\bm{\mathcal{G}}}_{ \sU})^\top\cdot\bm{w}_n\right) \bm{w}_n)\\
=&\frac{1}{T} \sum_{t=1}^T \sum_{n=M+1}^N \sum_{m=M+1}^N 
\left(\left(\bm{\mathcal{G}}_{t, \sU}-\overline{\bm{\mathcal{G}}}_{ \sU}\right)^\top\cdot \bm{w}_n\right) \bm{w}_n^\top\cdot\\&\bm{w}_m\left(\left(\bm{\mathcal{G}}_{t, \sU}-\overline{\bm{\mathcal{G}}}_{ \sU}\right)^\top \cdot\bm{w}_m\right) \\
=&
\frac{1}{T} \sum_{t=1}^T \sum_{n=M+1}^N\left(\bm{\mathcal{G}}_{t, \sU}-\overline{\bm{\mathcal{G}}}_{ \sU}\right)^T\cdot  \bm{w}_n\left(\bm{\mathcal{G}}_{t, \sU}-\overline{\bm{\mathcal{G}}}_{ \sU}\right)^\top\cdot \bm{w}_n \\
=&\frac{1}{T} \sum_{t=1}^T \sum_{n=M+1}^N \bm{w}_n^\top\cdot\left(\bm{\mathcal{G}}_{t, \sU}-\overline{\bm{\mathcal{G}}}_{\sU}\right)\left(\bm{\mathcal{G}}_{t, \sU}-\overline{\bm{\mathcal{G}}}_{ \sU}\right)^\top\cdot \bm{w}_n.
\end{aligned}
\end{equation}
Minimizing $\mathcal{J}$ is equivalent to reducing the variance of the pruned samples. Consequently, the goal outlined in \autoref{optimization matching} effectively becomes maximizing the variance of coreset shown in \autoref{variance}. 
\begin{figure*}
    \centering
    \includegraphics[width=0.8\linewidth]{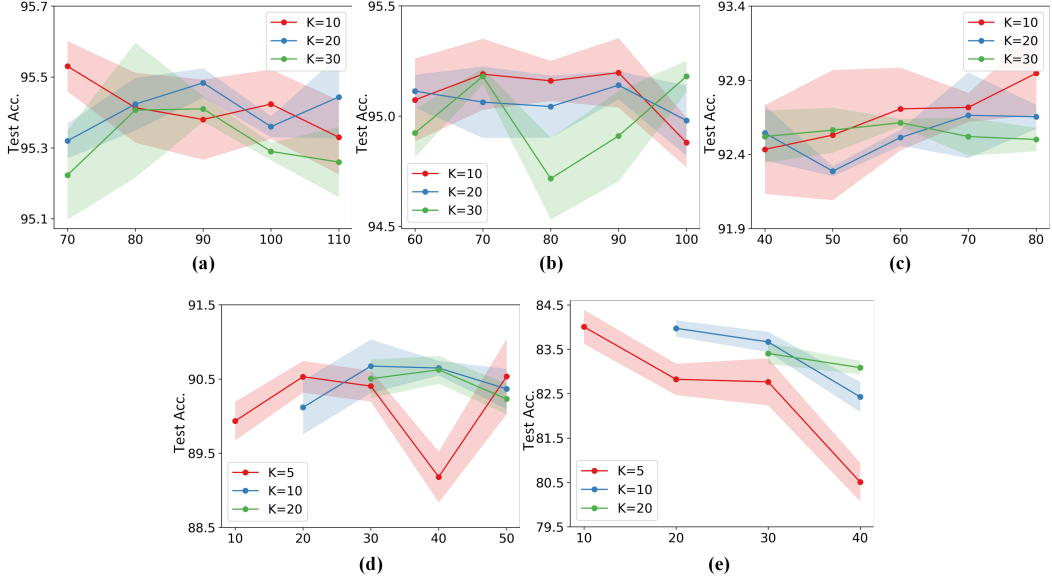}
    \caption{Parameter analysis of range $T$ and window size $K$ on CIFAR-10 with ResNet-18. From left to right, the corresponding pruning rates are $0.3$, $0.5$, $0.7$, $0.8$, and $0.9$.}
    \label{para-10}
\end{figure*}

\begin{table*}[h]
\renewcommand{\arraystretch}{0.95}
	\caption{Accuracy results on CIFAR-10 and 100 with smaller batch size. With a smaller batch size, all compared methods are enhanced under aggressive pruning, while our superiority remains consistent.}
	\label{smaller-bs-cifar10-100}
	\begin{center}
						\setlength{\tabcolsep}{15pt} 
\renewcommand{\arraystretch}{0.9} 
\resizebox{0.95\textwidth}{!}{
			\begin{tabular}{ccccccccc}
                   \toprule
                   &\multicolumn{4}{c}{\textbf{CIFAR-10}} &\multicolumn{4}{c}{\textbf{CIFAR-100}}\\ 
                   $p$& \multicolumn{2}{c}{80\%}& \multicolumn{2}{c}{90\%}& \multicolumn{2}{c}{80\%}& \multicolumn{2}{c}{90\%}\\ \cmidrule(r){1-1} \cmidrule(r){2-3} \cmidrule(r){4-5}\cmidrule(r){6-7}\cmidrule(r){8-9}
				batch size&\fformat{128}&\fformat{64}&\fformat{128}&\fformat{32}&\fformat{128}&\fformat{64}&\fformat{128}&\fformat{32}
\\ \cmidrule(r){1-1}\cmidrule(r){2-3} \cmidrule(r){4-5}\cmidrule(r){6-7}\cmidrule(r){8-9}
Random  & \format{86.92}{0.28}& \format{88.87}{0.47}&\format{76.71}{0.15}&\format{83.77}{0.27}& \format{56.19}{1.09}& \format{57.79}{0.24}&\format{34.88}{1.74}&\format{46.68}{1.07}\\
Entropy \cite{coleman2019selection}&\format{80.77}{0.26}&\format{83.49}{0.21}&\format{63.65}{0.62}&\format{72.06}{0.81}&\format{38.55}{1.49}&\format{42.86}{0.25}&\format{24.09}{0.47}&\format{29.56}{0.54}\\	
Forgetting \cite{toneva2018an}&\format{61.94}{1.33}&\format{76.18}{3.18}&\format{38.95}{0.28}&\format{45.87}{1.87}&\format{38.11}{0.55}&\format{38.42}{1.13}&\format{19.88}{0.69}&\format{25.82}{0.52}\\	
EL2N \cite{paul2021deep}& \format{59.28}{3.62}&\format{68.64}{3.70}&\format{23.54}{0.69}&\format{31.89}{1.51}& \format{14.67}{0.94}&\format{17.31}{0.33}&\format{5.54}{0.08}&\format{9.10}{0.69}\\
AUM \cite{pleiss2020identifying}& \format{59.11}{3.46} & \format{69.60}{3.11}&\format{30.62}{0.29}&\format{34.74}{0.11}& \format{16.85}{0.49} & \format{18.43}{0.47}&\format{7.99}{0.17}&\format{9.29}{0.27}\\
Moderate \cite{xia2022moderate}&\format{86.45}{0.31}&\format{87.76}{0.28}&\format{76.11}{2.25}&\format{83.61}{0.24}&\format{54.22}{0.58}&\format{56.52}{0.37}&\format{30.50}{1.21}&\format{41.82}{1.11}\\
Dyn-Unc \cite{he2023large}&\format{73.28}{0.50}&\format{79.76}{1.09}&\format{31.99}{0.74}&
\format{37.12}{1.12}&\format{36.21}{0.18}&\format{39.19}{0.27}&\format{11.68}{0.08}&\format{15.20}{0.41}\\
\textbf{TDDS}&\blformat{89.82}{0.15}&\blformat{91.30}{0.25}&\blformat{77.96}{0.29}&\blformat{85.46}{0.21}&\blformat{59.56}{0.42}&\blformat{63.01}{0.12}&\blformat{51.32}{0.16}&\blformat{54.51}{0.22}\\ \bottomrule    
\end{tabular}
}
\end{center}
\end{table*}
\begin{table*}[t]
\renewcommand{\arraystretch}{1.0}
	\caption{Cross-architecture generalization performance on CIFAR-10 with ResNet-18.}
	\label{cross-a-cifar10}
	\begin{center}
   			\setlength{\tabcolsep}{6pt} 
\renewcommand{\arraystretch}{1.2} 
\resizebox{1.0\textwidth}{!}{
			\begin{tabular}{ccccccccccccc}
				\toprule&\multicolumn{3}{c}{\textbf{ResNet-50}}
                    &\multicolumn{3}{c}{\textbf{VGG-16}}& \multicolumn{3}{c}{\textbf{MobileNet-v2}}& \multicolumn{3}{c}{\textbf{ShuffleNet}}\\ \cmidrule(r){2-4} \cmidrule(r){5-7}\cmidrule(r){8-10}\cmidrule(r){11-13}
$p$&30\%&\fformat{50\%}&\fformat{70\%}&30\%&\fformat{50\%}&\fformat{70\%}&30\%&\fformat{50\%}&\fformat{70\%}&30\%&\fformat{50\%}&\fformat{70\%}
\\ \cmidrule(r){1-1}\cmidrule(r){2-4} \cmidrule(r){5-7}\cmidrule(r){8-10}\cmidrule(r){11-13}
Random  & \fformat{94.33}&\fformat{93.40}&\fformat{90.94}&\fformat{92.93}&\fformat{91.52}&\fformat{88.55}&\fformat{91.83}&\fformat{91.69}&\fformat{89.66}&\fformat{90.86}&\fformat{89.08}&\fformat{85.87}\\
Entropy \cite{coleman2019selection}&\fformat{94.44}&\fformat{92.11}&\fformat{85.67}&\fformat{93.20}&\fformat{90.05}&\fformat{85.42}&\fformat{91.69}&\fformat{86.29}&\fformat{89.92}&\fformat{90.46}&\fformat{87.56}&\fformat{82.03}\\	
Forgetting \cite{toneva2018an}&\fformat{95.36}&\fformat{95.29}&\fformat{90.56}&\fformat{94.03}&\fformat{93.71}&\fformat{90.14}&\fformat{93.29}&\fformat{93.54}&\fformat{91.11}&\fformat{92.08}&\fformat{90.69}&\fformat{80.37}\\	
EL2N \cite{paul2021deep}&   \fformat{95.44}&\fformat{94.61}&\fformat{87.48}&   \fformat{93.86}&\fformat{93.19}&\fformat{87.23}&\fformat{92.96}&\fformat{92.99}&\fformat{88.38}&\fformat{92.12}&\fformat{90.73}&\fformat{79.63}\\
AUM \cite{pleiss2020identifying}& \fformat{95.07}&\fformat{95.26}&\fformat{91.36}& \fformat{94.14}&\fformat{93.73}&\fformat{88.44}&\fformat{93.43}&\fformat{93.37}&\fformat{90.97}&\fformat{92.23}&\fformat{91.72}&\fformat{79.41}\\
Moderate \cite{xia2022moderate}&\fformat{93.86}&\fformat{92.58}&\fformat{90.56} &\fformat{92.57}&\fformat{90.80}&\fformat{87.94}&\fformat{91.86}&\fformat{90.82}&\fformat{89.06}&\fformat{90.03}&\fformat{89.05}&\fformat{84.66}\\
Dyn-Unc \cite{he2023large}&\fformat{94.80}	&\fformat{94.21}	&\fformat{87.28}	&\fformat{92.98}	&\fformat{92.1}	&\fformat{86.99}&\fformat{92.16}	&\fformat{92.08}	&\fformat{89.93}				&\fformat{90.29}	&\fformat{88.80}	&\fformat{80.70}
\\
CCS \cite{zheng2022coverage}&\fformat{95.40}&\fformat{95.04}&\fformat{93.00} &\fformat{94.01}&\fformat{93.34}&\fformat{91.18}&\fformat{93.30}&\fformat{93.15}&\fformat{91.88}&\fformat{91.61}&\fformat{90.84}&\fformat{88.38}\\	 \textbf{TDDS}&\bblformat{95.50}&\bblformat{95.66}&\bblformat{93.92}&\bblformat{94.45}&\bblformat{93.74}&\bblformat{91.34}&\bblformat{94.52}&\bblformat{94.05}&\bblformat{92.31}&\bblformat{92.79}&\bblformat{92.07}&\bblformat{88.96}\\ \bottomrule
\end{tabular}
}
\end{center}
\end{table*}
\begin{table*}[h]
\renewcommand{\arraystretch}{1.2}
	\caption{Cross-architecture generalization performance on CIFAR-100 with ResNet-18.}
	\label{table: cross-a-cifar100-more}
	\begin{center}
						\setlength{\tabcolsep}{6pt} 
\renewcommand{\arraystretch}{1.2} 
\resizebox{0.99\textwidth}{!}{
			\begin{tabular}{ccccccccccccc}
				\toprule
                    &\multicolumn{3}{c}{\textbf{ResNet-50}}&\multicolumn{3}{c}{\textbf{VGG-16}}& \multicolumn{3}{c}{\textbf{MobileNet-v2}}& \multicolumn{3}{c}{\textbf{ShuffleNet}}\\ \cmidrule(r){2-4} \cmidrule(r){5-7}\cmidrule(r){8-10}\cmidrule(r){11-13}
$p$&30\%&\fformat{50\%}&\fformat{70\%}&30\%&\fformat{50\%}&\fformat{70\%}&30\%&\fformat{50\%}&\fformat{70\%}&30\%&\fformat{50\%}&\fformat{70\%}
\\ \cmidrule(r){1-1}\cmidrule(r){2-4} \cmidrule(r){5-7}\cmidrule(r){8-10}\cmidrule(r){11-13}
Random  & \fformat{72.09}&\fformat{68.27}&\fformat{61.75}&\fformat{71.75}&\fformat{67.57}&\fformat{61.03}&\fformat{70.27}&	\fformat{67.76}&	\fformat{63.02}&
\fformat{68.31}&	\fformat{65.17}&	\fformat{58.29}\\
Entropy \cite{coleman2019selection}&\fformat{73.09}&	\fformat{63.12}&	\fformat{47.61}& \fformat{69.52}&	\fformat{61.16}&	\fformat{48.42}& \fformat{67.91}&	\fformat{61.69}&	\fformat{51.74}& \fformat{64.12}&	\fformat{56.28}&	\fformat{44.68}\\	
Forgetting \cite{toneva2018an}&\fformat{78.17}&	\fformat{70.60}&	\fformat{48.74}&	\fformat{73.29}&	\fformat{66.01}&	\fformat{47.85}&	\fformat{72.37}&	\fformat{68.05}&	\fformat{54.06}& \fformat{66.94}&	\fformat{60.64}&	\fformat{40.65}
\\
EL2N \cite{paul2021deep}& \fformat{76.27}&	\fformat{65.83}&	\fformat{23.35}&	\fformat{72.42}&	\fformat{63.07}&	\fformat{36.47}&	\fformat{71.96}&	\fformat{63.81}&	\fformat{42.47}& \fformat{69.21}&	\fformat{56.82}&	\fformat{29.22}\\
AUM \cite{pleiss2020identifying}& \fformat{77.38}&	\fformat{64.2	}&\fformat{32.36}&	\fformat{73.60}&\fformat{62.01}&	\fformat{30.88}&	\fformat{72.29}&	\fformat{64.33}&	\fformat{36.35}& \fformat{66.98}&	\fformat{54.31}&	\fformat{29.24}\\
Moderate \cite{xia2022moderate}& \fformat{72.67}&	\fformat{68.75}&	\fformat{57.61}&	\fformat{70.1	}&\fformat{65.56}&	\fformat{57.80}&\fformat{70.01}&	\fformat{67.03}&	\fformat{60.78}& \fformat{66.53}&	\fformat{62.53}&	\fformat{50.33}\\
Dyn-Unc \cite{he2023large}&\fformat{73.00}&\fformat{63.7}	&\fformat{46.26}	&\fformat{68.48}	&\fformat{61.27}	&\fformat{47.24}	&\fformat{68.01}	&\fformat{62.04}&\fformat{49.40}	&\fformat{64.57}	&\fformat{56.14}	&\fformat{42.35}
\\
CCS \cite{zheng2022coverage} &\fformat{76.96}&	\fformat{72.43}&	\fformat{64.74}&	\fformat{74.02}&	\fformat{70.14}&	\bblformat{64.40}&\fformat{73.04}&	\fformat{70.63}&	\fformat{66.31}& \fformat{69.80}&\fformat{66.71}&	\bblformat{61.31}\\	
\textbf{TDDS}&\bblformat{79.53}&\bblformat{76.24}&\bblformat{66.56}&\bblformat{74.23}&\bblformat{70.66}&\fformat{64.08}&\bblformat{74.23}&\bblformat{71.14}&\bblformat{66.36}&\bblformat{70.14}&\bblformat{67.14}&\fformat{61.07}\\ \bottomrule
\end{tabular}
}
\end{center}
\end{table*}
\begin{figure*}[h]
    \centering
    \includegraphics[width=1\linewidth]{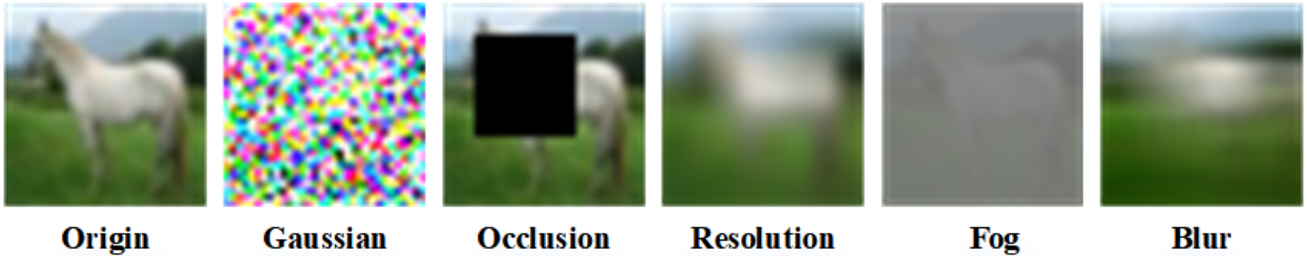}
    \caption{Illustration of the different types of noise used for image corruption. Here we consider Gaussian noise, random occlusion, resolution, fog, and motion blur.}
    \label{fig: robustness}
\end{figure*}
\section{Comparision Methods}\label{baselines}
\textbf{Random} randomly selects partial data from the full dataset to form a coreset.\\
\textbf{Entropy} \cite{coleman2019selection} is a metric of sample uncertainty. Samples with higher entropy are considered to have a greater impact on model optimization. The entropy is calculated with the predicted probabilities at the end of training. \\
\textbf{Forgetting} \cite{toneva2018an} counts how many times the forgetting happens during the training. The unforgettable samples can be removed with minimal performance drop.\\
\textbf{EL2N} \cite{paul2021deep} selects samples with larger gradient magnitudes which can be approximated by error vector scores. Only the first $10$-epoch error vector scores are averaged to evaluate samples.\\
\textbf{AUM} \cite{pleiss2020identifying} selects samples with the highest area under the margin, which measures the probability gap between the target class and the next largest class across all training epochs. A larger AUM suggests higher importance.\\
\textbf{Moderate} \cite{xia2022moderate} calculates sample-wise distance in feature space. Samples near the median are considered more important. Here, the features are generated by a pretrained model.\\
\textbf{CCS} \cite{zheng2022coverage} uses a variation of stratified sampling across importance scores to improve the coverage of coreset, which can be combined with other criteria. In our experiments, AUM~\cite{pleiss2020identifying} is used as the importance measurement in CCS~\cite{zheng2022coverage}.\\
\textbf{Dyn-Unc} \cite{he2023large} calculates the dynamic uncertainty defined as the variance of target class predicted probabilities during the training progress. The samples with larger uncertainties are more important than those with smaller uncertainties.
\\Note that, we use the same experimental hyperparameter settings to ensure equity for all the compared methods. 
\section{Parameter Settings}\label{parameter-settings}
The grid search of CIFAR-10 is shown in \autoref{para-10}. The optimal ($p, T, K$) setting is $(0.3, 70, 10)$, $(0.5, 90, 10)$, $(0.7, 80, 10)$, $(0.8, 30, 10)$, and $(0.9, 10, 5)$. For ImageNet-1K with ResNet-34, we set $(0.3, 20, 10)$, $(0.5, 20, 10)$, and $(0.7, 30, 20)$. For ImageNet-1K with Swin-T, we set $(0.25, 200, 10)$, $(0.3, 180, 10)$, $(0.4, 150, 10)$, and $(0.5, 100, 10)$. 
\section{Small Batch Size Boosts Aggressive Pruning}\label{smaler-bs-at-high-pruning-rate}
In the experiments, we reveal that smaller batch size boosts coreset training, especially under aggressive pruning rates. This phenomenon is attributed to so-called \textit{Generalization Gap} \cite{keskar2016large}, which suggests that when the available data is extremely scare, smaller batch size can prevent overfitting by allowing more random explorations in the optimization space before converging to an optimal minimum.
As reported in \autoref{smaller-bs-cifar10-100}, smaller batch size improves the accuracy of high pruning rates for all the compared methods. Note that, regardless of the batch size, our method consistently demonstrates a significant advantage.
\section{Generalization Across Architectures}\label{across_more}

We conduct cross-architecture experiments to examine whether coresets perform well when being selected on one architecture and then tested on other architectures. Four representative architectures including ResNet-50, VGG-16, MobileNet-v2, and ShuffleNet are used to assess the cross-architecture generlization.
\autoref{cross-a-cifar10} lists the results on CIFAR-10, while \autoref{table: cross-a-cifar100-more} reports the results on CIFAR-100. We can see the coresets constructed by the proposed TDDS achieves stably good testing results, regardless of which model architecture is used to test. Experiments on CIFAR-100 are reported in Supplementary.

\begin{table*}[t]
\renewcommand{\arraystretch}{0.9}
	\caption{Robustness to image corruption on CIFAR-100 with ResNet-18. 20\% training images are corrupted. The model trained with the full dataset achieves \(75.30\%\) accuracy.}
	\label{table: robustness}
	\begin{center}
						\setlength{\tabcolsep}{15pt} 
\renewcommand{\arraystretch}{0.8} 
\resizebox{0.8\textwidth}{!}{
			\begin{tabular}{ccccccc}
				\toprule
                    & \multicolumn{5}{c}{\textbf{Image Corruption}}\\ 
$p$&\fformat{30\%}&\fformat{50\%}&\fformat{70\%}&80\%&\fformat{90\%}
\\ \cmidrule(r){1-1}\cmidrule(r){2-6} 
Random  &\format{71.34}{0.29}&\format{67.17}{0.43}&\format{58.56}{0.87}&\format{51.85}{0.48}&\format{37.30}{0.60}
\\Entropy \cite{coleman2019selection}    &\format{68.83}{0.17}&\format{62.19}{0.26}&\format{49.25}{0.20}&\format{41.26}{0.24} &\format{28.03}{0.44}\\	
Forgetting \cite{toneva2018an}    &\format{73.77
}{0.07}&\format{65.38}{1.87}&\format{47.41}{1.11}&\format{36.07}{1.44}&\format{22.02}{0.40}\\		
EL2N \cite{paul2021deep}     &\format{70.58}{0.30}&\format{48.17}{3.26}&\format{14.47}{0.73}&\format{12.21}{0.35}&\format{8.62}{0.29}\\
AUM \cite{pleiss2020identifying}    &\format{71.14}{0.63}&\format{44.06}{1.80}&\format{13.89}{0.43}&\format{8.06}{0.15}&\format{4.93}{0.15}
\\
Moderate \cite{xia2022moderate}    &\format{72.20}{0.11} &\format{67.52}{0.18}&\format{59.84}{0.17}&\format{52.89}{0.12}&\format{36.16}{1.23} \\
Dyn-Unc \cite{he2023large}&\format{67.74}{0.38}&\format{59.40}{0.15}&\format{45.39}{0.37}&\format{34.11}{0.47}&\format{13.55}{0.29}
\\
CCS \cite{zheng2022coverage}&\format{70.14}{0.14}&\format{64.77}{0.31}&\format{54.95}{1.08}&\format{44.95}{0.69}&\format{30.16}{1.13}
\\
TDDS&\blformat{75.40}{0.12}&\blformat{72.49}{0.22}&\blformat{65.84}{0.30}&\blformat{60.85}{0.07}&\blformat{49.35}{0.12}\\
				\bottomrule
			\end{tabular}
   }
  \end{center}
\end{table*}
\begin{table*}[t]

	\caption{Robustness to label noise on CIFAR-100 with ResNet-18. 20\% training samples are mislabeled. The model trained with the full dataset achieves \(65.48\%\) accuracy.}
	\label{table: robustness_1}
	\begin{center}
						\setlength{\tabcolsep}{15pt} 
\renewcommand{\arraystretch}{0.8} 
\resizebox{0.8\textwidth}{!}{
			\begin{tabular}{ccccccc}
				\toprule
                    & \multicolumn{5}{c}{\textbf{Lable Noise}}\\ 
$p$&\fformat{30\%}&\fformat{50\%}&\fformat{70\%}&80\%&\fformat{90\%}
\\ \cmidrule(r){1-1}\cmidrule(r){2-6} 
Random  &\format{62.17}{0.42}&\format{55.3}{0.25}&\format{40.8}{0.49}&\format{34.41}{0.69}&\format{22.74}{0.27}
\\
Entropy \cite{coleman2019selection}    &\format{60.01}{0.59}&\format{54.27}{0.90}&\format{42.75}{0.80}&\format{35.18}{0.28}&\format{24.34}{1.01}
\\	
Forgetting \cite{toneva2018an}    &\format{58.75}{0.28}	&\format{47.90}{0.79}&\format{29.34}{0.51}&\format{21.38}{0.34}&\format{13.31}{0.35}
\\		
EL2N \cite{paul2021deep}     &\format{63.76}{0.07}&\format{50.39}{0.89}&\format{20.89}{1.79}&\format{10.20}{0.95}&\format{5.97}{0.19}
\\
AUM \cite{pleiss2020identifying}    &\format{50.49}{0.81}&\format{22.86}{0.11}&\format{5.79}{0.36}&\format{2.31}{0.39}&\format{1.25}{0.04}
 \\
Moderate \cite{xia2022moderate} &\format{61.58}{0.29}&\format{57.23}{0.05}&\format{49.28}{0.25}&\format{43.25}{1.02}&\format{32.07}{0.25} 	\\
Dyn-Unc \cite{he2023large}&\format{52.99}{0.34}&\format{38.83}{0.17}&\format{19.17}{0.15}&\format{3.41}{0.04}&\format{1.64}{0.08}
\\
CCS \cite{zheng2022coverage}&\format{53.38}{0.86}&\format{40.59}{0.21}&\format{25.30}{0.17}&\format{20.49}{0.43}&\format{15.49}{0.61}
\\
TDDS&\blformat{65.15}{0.06}&\blformat{62.72}{0.37}&\blformat{54.97}{0.20}&\blformat{50.14}{0.20}&\blformat{39.32}{0.19}\\
				\bottomrule
			\end{tabular}
   }
  \end{center}
\end{table*}
\section{Robustness to Complex Realistic Scenarios}\label{robustness}
We also investigate the robustness of corsets in complex and realistic scenarios, including image corruption and label noise. Following the settings stated in \cite{xia2022moderate}, we consider five types of realistic noise, namely Gaussian noise, random occlusion, resolution, fog, and motion blur (shown in \autoref{fig: robustness}). Here, the ratio for each type of corruption is 4\%, resulting in a total 20\% of training images being corrupted. Besides, we also consider label noise by replacing the original label with labels from other classes. The mislabel ratio is also set to 20\%. The results reported in \autoref{table: robustness} and \autoref{table: robustness_1} verify the robustness of our proposed TDDS in complex and realistic scenarios.

\end{document}